\documentclass[10pt,twocolumn,letterpaper]{article}

\usepackage{iccv}
\usepackage{appendix}
\usepackage{times}
\usepackage{epsfig}
\usepackage{graphicx}
\usepackage{amsmath}
\usepackage{amssymb}
\usepackage[ruled,vlined]{algorithm2e}
\usepackage{multirow}

\iccvfinalcopy 
\usepackage{xcolor, soul}
\usepackage{amsfonts}
\usepackage{dsfont}
\usepackage{authblk}

\usepackage[pagebackref=true,breaklinks=true,colorlinks,bookmarks=false]{hyperref}

\def\iccvPaperID{8009} 

\begin{document}

\title{Learning of Visual Relations: The Devil is in the Tails}


\author[1]{Alakh Desai\thanks{Authors have equal contributions.}}
\newcommand\CoAuthorMark{\footnotemark[\arabic{footnote}]} 
\author[1]{Tz-Ying Wu\protect\CoAuthorMark} 
\author[2]{Subarna Tripathi}
\author[1]{Nuno Vasconcelos}
\affil[1]{University of California San Diego, USA}
\affil[2]{Intel Labs, USA}
\renewcommand\Authands{, and }

\maketitle
\ificcvfinal\thispagestyle{empty}\fi

\begin{abstract}
Significant effort has been recently devoted to modeling visual relations.
This has mostly addressed the design of architectures, typically by adding parameters and increasing model complexity.
However, visual relation learning is a long-tailed problem, due to the combinatorial nature of joint reasoning about groups of objects.
Increasing model complexity is, in general, ill-suited for long-tailed problems due to their tendency to overfit. 
In this paper, we explore an alternative hypothesis, denoted {\bf the Devil is in the Tails}.
Under this hypothesis, better performance is achieved by keeping the model simple but improving its ability to cope with long-tailed distributions. 
To test this hypothesis, we devise a new approach for training visual relationships models, which is inspired by state-of-the-art long-tailed recognition literature.
This is based on an iterative decoupled training scheme, denoted Decoupled Training for Devil in the Tails (DT2).
DT2 employs a novel sampling approach, Alternating Class-Balanced Sampling (ACBS), to capture the interplay between the long-tailed entity and predicate distributions of visual relations.
Results show that, with an extremely simple architecture, DT2-ACBS significantly outperforms much more complex state-of-the-art methods on scene graph generation tasks.
This suggests that the development of sophisticated models must be considered in tandem with the long-tailed nature of the problem.
\end{abstract}
\label{abs}

\section{Introduction}
\label{sec:intro}

\begin{figure}
    \centering
    \includegraphics[width=0.4\textwidth]{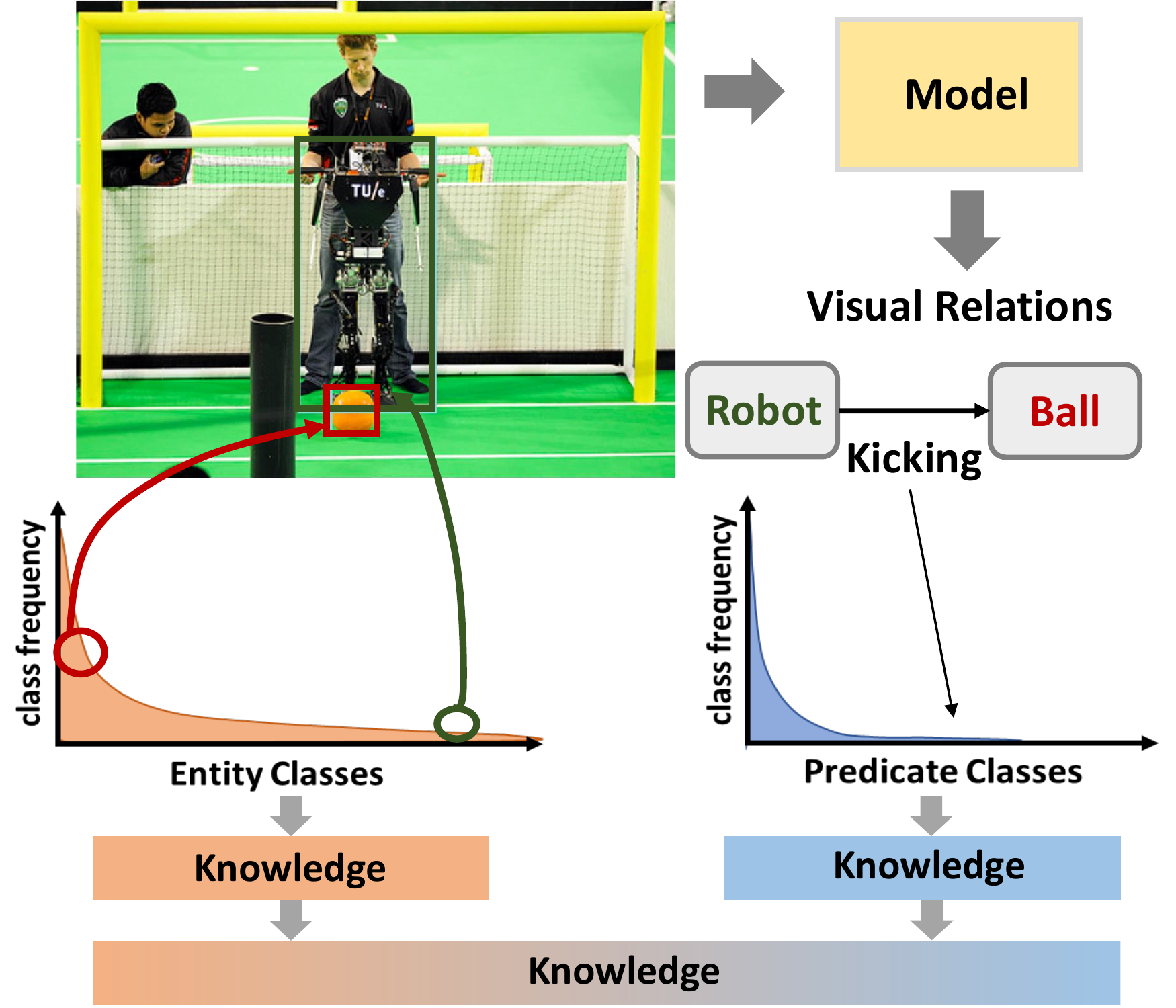}
    \caption{The devil is in the tails: Architecture design and learning process of visual relations need to consider the long-tailed nature of both entity and predicate class distributions.}
    \label{fig:teaser}
\end{figure}

Scene graphs provide a compact structured description of complex scenes and the semantic relationships between objects/entities. Modeling and learning such visual relations benefit several high-level Vision-and-Language tasks such as caption generation \cite{For_image_captioning_Yao_2018_ECCV, SC_autoencoding_Yang_2019_CVPR}, visual question answering \cite{GQA_Hudson_2019_CVPR}, image retrieval \cite{SG_for_IR_15, retrieval_Wang0YSC20}, image generation \cite{SG2im_Johnson_2018_CVPR, pastegan_LiMBDWW19, retrievegan2020} and robotic manipulation planning \cite{20-oz-CoRL}. 
Scene graph generation requires the understanding of the locations and the class associated with the entity as well as the relationship between a pair of entities. The relationship between a pair of entities is usually formulated as a \textit{$<subject-predicate-object>$} tuple, where subject and object are two entities. 
Scene graph generation (SGG) faces the challenges from both the long-tailed entity recognition problem and visual relation recognition problem. 

While long-tailed entity recognition has been addressed in the literature~\cite{oltr,cao2019learning,cbloss,iclr2020}, the imbalance becomes more prevalent for the SGG tasks, owing to the severe long-tailed nature of the predicate distribution.
Take Figure~\ref{fig:teaser} for example. While the class of the subject (``ball") is popular, the class of the object (``robot") and the predicate (``kicking") can be infrequent, leading to the rare occurrence of the tuple ``robot-kicking-ball". This shows that even when the entity class distribution is balanced, the imbalanced predicate class distribution can lead to a more imbalanced tuple distribution. Of course, such imbalance issues can be exacerbated if both entity classes and predicate classes are skewed (e.g. ``tripod-mounted-on-donkey"). The combination of long-tailed entity and predicate classes makes SGG a more challenging problem. 

While the long-tailed problem poses a great challenge to SGG tasks, it has not been well addressed in the SGG literature. Existing works~\cite{neural_motifs_2017, graph_RCNN_Yang_2018_ECCV, kern_CVPR19, VCTree_Tang_2019_CVPR, graphical_contrastive_loss19} instead focused on designing more complex models, primarily by adding architectural enhancements that increase model size. While this has enabled encouraging performance under the Recall@k (R@k) metric, this metric is biased toward the highly populated classes. This suggests that prior works may be overfitting on popular predicate classes (e.g. \emph{on/has}), but their performances could degrade on the less frequent classes (e.g. \emph{eating/riding}). Such a bias towards the populated classes is problematic, because predicates lying in the tails often provide more informative depictions of scene content. 
The failure to predict tail classes could lead to a less informative scene graph 
, limiting the effectiveness of scene graphs for intended applications.
In this paper, we explore the hypothesis that the Devil is in the tails. Under this hypothesis, visual relation learning is better addressed by a simple model of improved ability to cope with long-tailed distributions. 

To investigate this hypothesis, we first analyze the distribution of entity and predicate classes in the Visual Genome dataset. As shown in Figure \ref{fig:clsR_baseline}, both distributions are heavily skewed, but with different magnitude. The imbalance in the predicate distribution is more severe than that in the entity distribution. To the best of our knowledge, none of the existing SGG methods considered the jointly long-tailed distributions of entity and predicate classes. 
To address this, we propose a new approach to visual relationship learning, based on a simpler architecture than those in the literature but a more sophisticated training procedure, denoted {\it Decoupled Training for Devil in the Tails (DT2)}.

DT2 is a generalization of the decoupled training procedures that have recently become popular for long-tailed recognition~\cite{iclr2020}. It consists of an alternative sampling scheme that produces distributions balanced for entities and predicates. This is accompanied by a novel sampling scheme, {\it Alternating Class-Balanced Sampling (ACBS)},
which captures the interplay between the two different long-tailed distributions through an implementation of learning without forgetting~\cite{lwf2016} based on a mechanism that introduces memory between the sampling iterations, using knowledge distillation. 
With DT2, we show that a simple architecture with \textbf{10$\times$} fewer parameters significantly outperforms prior, and more sophisticated, architectures designed for SGG, under the mRecall@K metric, which is suited for measuring the performance of a long-tailed dataset. 
Ablation studies of different sampling schemes as well as analysis of performance on classes of different popularity further validate our hypothesis.

Overall, the paper makes three contributions.
1) We devise a simple model architecture with the decoupled training scheme, namely \textbf{DT2}, suited for the long-tailed SGG tasks.
2) We propose a novel sampling strategy, \textbf{Alternating Class-Balanced Sampling (ACBS)},
to capture the interplay between different long-tailed distributions of entities and relations.
3) The combined \textbf{DT2-ACBS} significantly outperforms state-of-the-art methods of more complex architectures on all SGG tasks on the Visual Genome benchmark. The code is available on the project website\footnote{\url{http://www.svcl.ucsd.edu/projects/DT2-ACBS}}.

\section{Related work}
\label{sec:related}
\subsection{Scene graph generation}

Several works have addressed the generation of scene graphs for images \cite{Zareian_2020_ECCV, PCPL_YanSJHJC020, Zareian_2020_CVPR, GIP2018, relational_embedding_NIPS2018, xu2017scenegraph, neural_motifs_2017, graph_RCNN_Yang_2018_ECCV, FactorizableNet_Li_2018_ECCV, Gu_2019_CVPR, kern_CVPR19, VCTree_Tang_2019_CVPR, UVTransE, graphical_contrastive_loss19, visual_rel_as_func_19}.
Most approaches focus on either sophisticated architecture design or contextual feature fusion strategies, such as message passing and recurrent neural networks \cite{neural_motifs_2017, VCTree_Tang_2019_CVPR}, to optimize SGG performance on the Visual Genome dataset \cite{visual_genome16} under the Recall@K metric.
While these approaches achieved gains for highly populated classes,  underrepresented classes tend to have much poorer performance. 
Recently, \cite{kern_CVPR19,TDETangNHSZ20, PCPL_YanSJHJC020,wen2020unbiased,li2021bgnn} started to address the learning bias induced by the dataset statistics, by using a more suitable evaluation metric, mRecall@K, which averages recall values across classes. To address the dataset bias, TDE \cite{TDETangNHSZ20} employed causal inference in the prediction stage , whereas \cite{wen2020unbiased} used a pseudo-siamese network to extract balanced visual features, and PCPL~\cite{PCPL_YanSJHJC020} harnessed implicit correlations among predicate classes and used a complex graph encoding module consisting of a number of stacked encoders and attention heads. A concurrent work \cite{li2021bgnn} introduces confidence-based gating with bi-level data resampling to mitigate the training bias.
These methods considered, at most, the long-tailed distribution of either predicates or entities and do not disentangle the gains of sampling from those of complex architectures. For example, \cite{PCPL_YanSJHJC020} proposed a contextual feature generator via graph encoding with 6 stacked encoders, each with 12 attention heads and a feed-forward network. We argue that long-tailed distributions should be considered for both entities and predicates and show that, when this is done, better results can be achieved with a much simpler architecture.

\subsection{Long-tailed recognition}\label{sec:long-tail_related}
Prior work addresses the long-tailed issue in 3 directions: data re-sampling, cost-sensitive loss and transfer learning.

\textbf{Data resampling}~\cite{5128907, 4633969, Zou_2018_ECCV, Han2005, Drummond2003C4, Chawla2002} is a popular strategy to oversample tail (underrepresented) classes and undersample head (populated) classes. Oversampling is achieved either by duplicating samples or by synthesizing data~\cite{4633969, Zou_2018_ECCV, Chawla2002}. While producing a more uniform training distribution, recent works~\cite{iclr2020, zhou2020BBN} argue that this strategy is unsuitable for deep representation learning like CNN. \cite{iclr2020} decouples the representation learning from the classifier learning, adopting different sampling strategies in the two stages, whereas
\cite{zhou2020BBN} proposes a two-stream model with a mixed sampling strategy.
The proposed method lies in this direction, since we consider different distributions of entity and predicate classes, and adopt different sampling strategies for training different model components.

\textbf{Cost-sensitive losses}~\cite{Qi17, cbloss, cao2019learning, FocalLoss} assign different costs to the incorrect prediction of different samples, according to class frequency~\cite{cbloss, cao2019learning} or difficulty~\cite{Qi17, FocalLoss}. This is implemented by assigning higher weights or enforcing larger margins for classes with fewer samples. Weights can be proportional to inverse class frequency or effective number~\cite{cbloss} and can be estimated by meta-learning \cite{Jamal_2020_CVPR}. This re-weighting strategy was recently applied to the scene graph literature \cite{PCPL_YanSJHJC020} to overcome long-tailed distributions.

\textbf{Transfer learning} methods transfer information from head to tail classes.
\cite{Wang2016LearningTL, NIPS2017_7278} learns to predict few-shot model parameters from many-shot model parameters, and ~\cite{oltr} proposes a meta-memory for knowledge sharing. \cite{Wu20DeepRTC} leverages a hierarchical classifier to share knowledge among classes. \cite{Xiang2020LearningFM} learn an expert model for each class popularity, and combine them by knowledge distillation. 

\section{Formulation and data statistics}
\label{sec:approach}
In this section, we review the problem of learning visual relations and discuss its long-tailed nature.

\subsection{Definitions}
The inference of the visual relationships in a scene is usually formulated as a three stage process. The objects/entities in the scene are detected, classified, and the relationships between each pair of entities, in the form of predicates, are finally inferred. \cite{SG_for_IR_15} formulated these stages with a {\it Scene Graph}. Let $C$ and $P$ be the set of entity and predicate classes, respectively. Each entity $e=(e^b, e^c) \in \mathcal{E}$ is composed by a bounding box $e^b\in\mathbb{R}^4$ and a class label $e^c\in C$. A relation $r=(s,p,o)$ is a three-tuple, connecting a subject $s$ and an object $o$ identities ($s,o \in \mathcal{E}$), through a predicate $p \in P$. For example, {\it person-riding-bike}.
The scene graph $G=(E,R)$ of an image $I$ contains a set of entities $E=\{e_i\}_{i=1}^m$ and a set of relations $R=\{r_j\}_{j=1}^n$ extracted from the image. This can be further decomposed into a set of bounding boxes $B=\{e^b_i\}_{i=1}^m$, a set of class labels $Y=\{e^c_i\}_{i=1}^m$, and a set of relations $R$.

The generation of a scene graph $G$ from an image $I$ is naturally mapped into the probabilistic model
\begin{align}
Pr(G|I)=Pr(B|I)Pr(Y|B,I)Pr(R|B,Y,I),
\end{align}
where $Pr(B|I)$ is a bounding box prediction model, $Pr(Y|B,I)$ an entity class model and $Pr(R|B,Y,I)$ is a predicate class model. Joint inference of the three tasks is referred to as \textbf{Scene Graph Detection (SGDet)}. However, because bounding box prediction has been widely studied in object detection~\cite{faster-rcnn}, it is possible to simply adopt an off-the-shelf detector. This motivates two other tasks: \textbf{Predicate classification (PredCls)}, where both bounding boxes and entity classes are given, and \textbf{Scene Graph Classification (SGCls)}, where only bounding boxes are known. 

\subsection{Long-tailed visual relations}
Long-tailed distributions are a staple of the natural world, where different classes occur with very different frequencies. For example, while some entity classes (e.g. chair) occur very frequently, others (e.g donkey) are much less frequent.
Long tails are problematic because, under standard loss functions and evaluation metrics, they encourage machine learning systems to overfit on a few head classes and ignore a large number of tail classes. Recent works~\cite{oltr,cbloss,zhou2020BBN,iclr2020} have shown that sampling techniques which de-emphasize popular classes, giving more weight to rare ones, can induce very large recognition gains when distributions are long-tailed. However, the issue has not been thoroughly considered in the visual relations literature. 

\begin{figure}[t!]
    \centering
       \includegraphics[width=0.47\linewidth,height=0.25\linewidth]{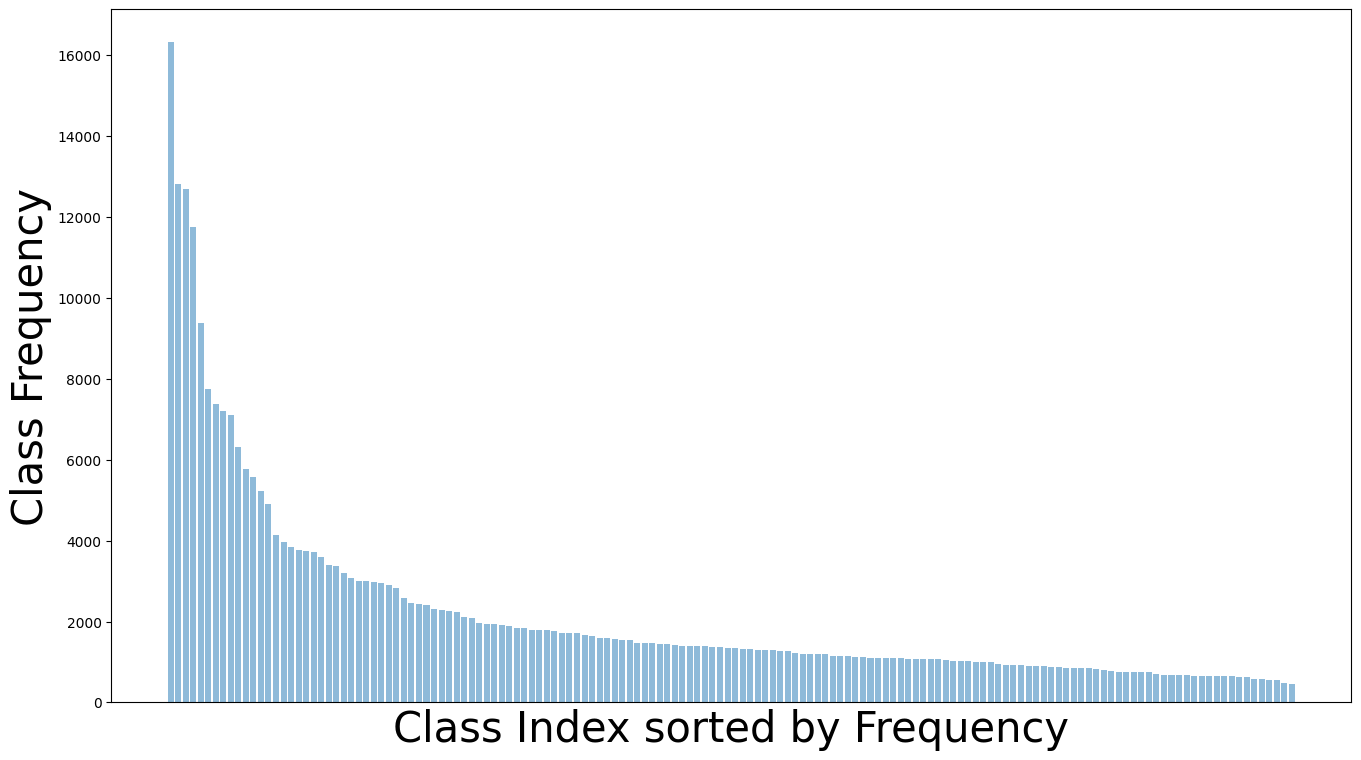}
       \includegraphics[width=0.47\linewidth,height=0.25\linewidth]{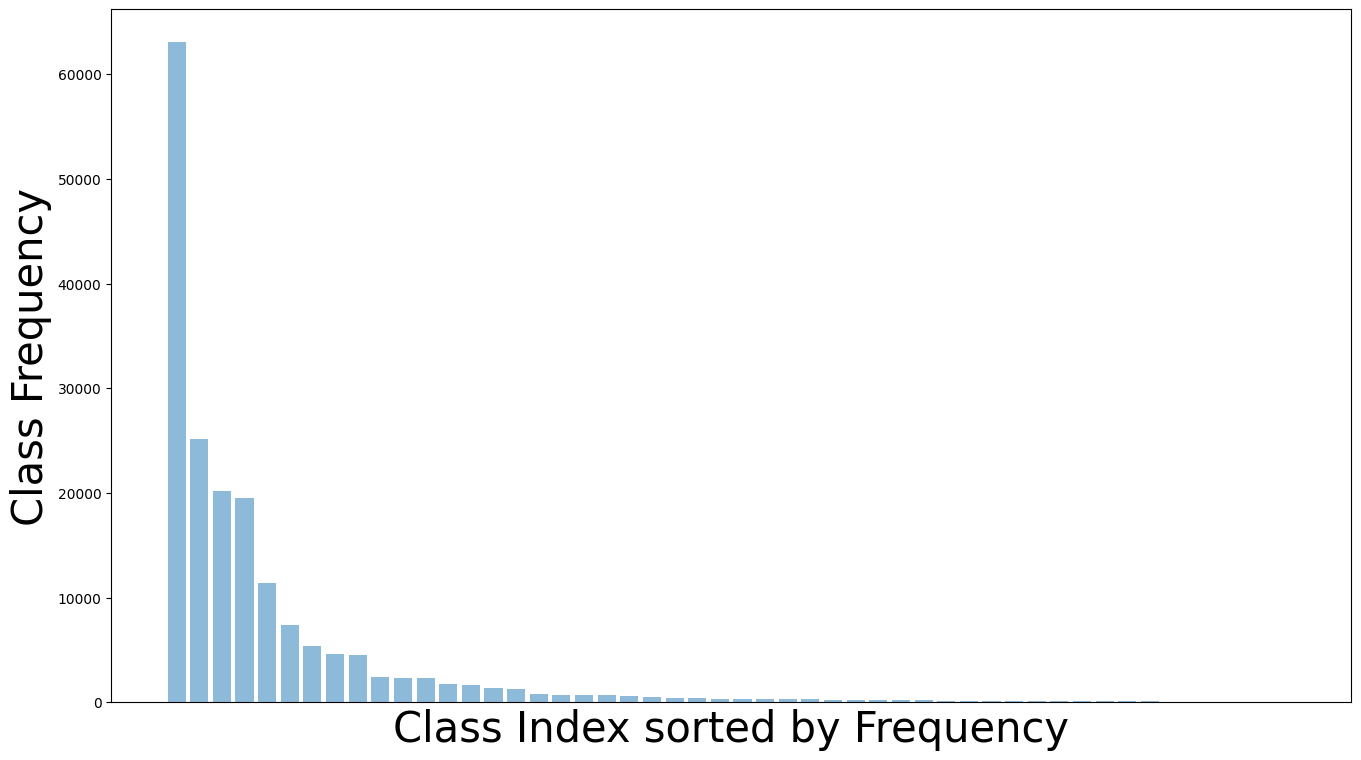}
    \caption{Object classes (left) and predicate classes (right) are both long-tailed distributed in Visual Genome (VG150).}
    \label{fig:clsR_baseline}
\end{figure}

This is somewhat surprising, given the combinatorial dependence of visual relationships on entities and predicates. Since entities are long-tailed, relationships between pairs of entities have even more skewed distributions.
For example, because the entity classes ``donkey" and ``cliff" are less frequent than ``chair" and ``leg", the relation ``donkey-on-cliff" is much less frequent than ``chair-has-leg". 
This, however, is not the only source of skew, since predicates can be rare even when associated entity classes are popular, e.g. \emph{playing} is much less popular than \emph{has}.
Finally, relationships can be rare even when involving frequent entities and predicates, e.g. the relation ``car-has-wheel" is much more likely than ``car-has-camera". For all these reasons, very long tails are unavoidable for visual relations. This is quite visible 
in the widely used Visual Genome~\cite{visual_genome16} dataset. As shown in Figure~\ref{fig:clsR_baseline}, both the distribution of entity and predicate classes are long-tailed. For entities, the most populated class is ${35\times}$ larger than the least populated. For predicates, the former is ${12,000\times}$ larger than the latter (${5,000\times}$ if the least frequent predicate class 
is discarded). 
Note that this is much larger than the square of the ratio between entity classes ($1,225$) suggested by the factorial nature of relationships. 

The long-tailed problem is exacerbated by the evaluation protocol, based on the Recall@K (R@K) measure, adopted in most of the
scene graphs literature. This measures the average percentage of ground truth relation triplets that appear in the top $K$ predictions and, like any average, is dominated by the most frequent relationship classes. Hence, it does not penalize solutions that simply ignore infrequent relationship classes.
Since most works, e.g.~\cite{VCTree_Tang_2019_CVPR, visual_rel_as_func_19, kern_CVPR19}, focus on designing ever more complex network architectures to optimize R@K performance, it is unclear whether all that is being accomplished is stronger overfitting to a few dominant classes (e.g. ``on"). This is undesirable for two reasons. First, the number of infrequent relations is much larger than that of dominant relationships. Second, while dominant relations include many obvious contextual relationships (e.g.``car-has-wheels"), infrequent ones are 
potentially more informative (e.g. ``monkey-playing-ball") of the scene content. In summary, the focus on optimizing R@K could lead to systems that are only capable of detecting a few relationships of relatively low information content.

This problem has been recognized in the recent literature, where some works~\cite{kern_CVPR19, TDETangNHSZ20} have started to adopt the mRecall@K (mR@K) metric, which first averages the recall of triplets within the same predicate class and then averages the class recalls over all the predicate classes. 
While this is a step in the right direction, it is not sufficient to account for class imbalance \emph{only} at the evaluation stage. Instead, the learning algorithm should explicitly address this imbalance. This leads to an alternative hypothesis that we explore in this work: {\it Is the devil in the tails?\/} Or, in other words, can a simple model designed explicitly to cope with the long-tailed nature of visual relations outperform existing models, which are much more complex but ignore this property? To investigate this hypothesis, we introduce a solution that uses a model much simpler than recently proposed architectures, but is much more sophisticated in its use of sampling techniques that target the long-tailed nature of visual relationship.
\begin{figure*}
    \centering
    \includegraphics[width=0.9\linewidth]{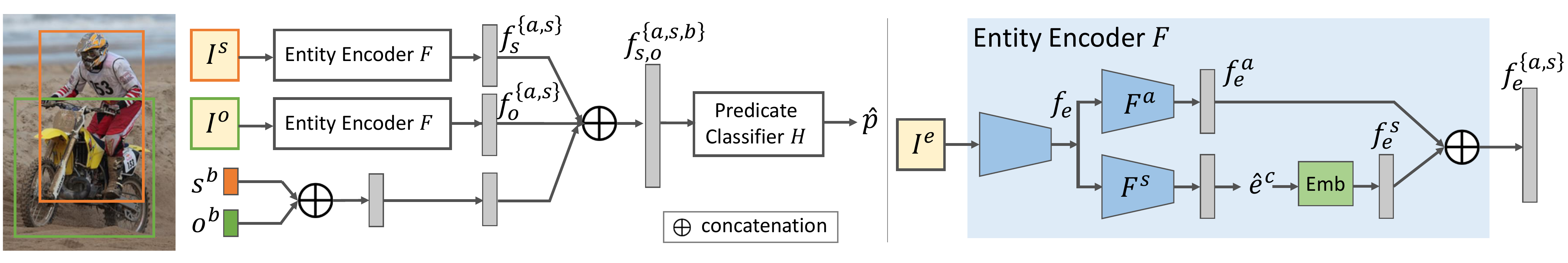}
    \caption{The model architecture of DT2 is composed of an entity encoder $F$ (right) and a predicate classifier $H$.}
    \label{fig:model}
\end{figure*}

\section{Method}
In this section, we introduce the proposed network architecture, losses, and the training procedure. 

\subsection{Notations}
For a relation tuple $r_j=(s_j,p_j,o_j)$ in image $I$, $p_j$ is the ground truth predicate class, while $s_j = (s_j^b,s_j^c)$ and $o_j = (o_j^b, o_j^c)$ are the subject and object entities, composed of its associated bounding box coordinates (e.g. $s_j^b$) and ground truth entity class (e.g. $s_j^c$). The bounding boxes of an entity can be either the ground truth coordinates or the predictions from a detection model, depending on the task of interest (i.e. SGCls or SGDet). With the bounding boxes, the corresponding image patch $I_j^s$ and $I_j^o$ for the subject and object can be cropped from the image $I$.

In addition, we define $\rho$ as a probability vector at the output of the softmax function with temperature $\tau$, and its $i^{th}$ entry is formulated as
\begin{align}
    \rho_i(f, {\bf W}, \tau) = \frac{\exp{(\mathbf{w}_i^T f / \tau)}}{\sum_{k} \exp{(\mathbf{w}_k^T f / \tau)}}~,\label{eq:softmax_formulation}
\end{align}
where $f\in\mathcal{R}^d$ is a feature vector, ${\bf W}\in \mathcal{R}^{d \times k}$ is the matrix of $k$ weight parameters $\mathbf{w}_k\in\mathcal{R}^d$.  

\subsection{Model architecture}\label{sec:model}
Figure~\ref{fig:model} summarizes the architecture of the {\it Decoupled Training for Devil in the Tails\/} (DT2) model. This combines an entity encoder $F$, as shown in the right part of Figure~\ref{fig:model}, and a predicate classifier $H$. DT2 takes the bounding box coordinates $s_j^b$, $o_j^b$~\cite{Chen_2019_ICCV}
and the corresponding cropped image patches $I_j^s$ and $I_j^o$ as input. The entity encoder $F$ is then applied to both $I_j^s$ and $I_j^o$, to extract a pair of subject-object feature vectors $f_s^{\{a,s\}}, f_o^{\{a,s\}}$ that represent both the \textit{appearance} and \textit{semantics} of entities $s_j$ and $o_j$. These are then concatenated with an embedding of the bounding box coordinates $s_j^b$ and $o_j^b$, and fed to a predicate classifier $H$. Implementation details of the entity encoder and the predicate classifier are elaborated below.

\textbf{Entity encoder} $F$ first maps image patch $I^e$ of entity $e$ through a feature extractor, implemented with the first three convolutional blocks of a pretrained ResNet101~\cite{resnet}. We use a faster R-CNN pre-trained for object detection on Visual Genome under regular sampling (all images are sampled uniformly). The resulting feature vector $f_e$ is then mapped to two feature vectors, $f_e^s$ and $f_e^a$, that encode semantics and appearance information respectively, through two different branches sharing identical architecture. The semantic branch $F^s(\cdot;\theta)$ of parameter $\theta$ is implemented with a stack of convolution layers (the last convolutional block of ResNet101). Its output is then fed to a softmax layer that predicts the 
probability 
$\bar{e}^c \in [0,1]^C$ of the class of the entity $e$, i.e.
\begin{align}
     \bar{e}^c = \rho(F^s(f_e;\theta), {\bf W}^e, \tau=1)~,\label{eq:softmax}
 \end{align}
where ${\bf W}^e$ is the matrix of the entity classifier weights and $\tau$ of $\rho$ in (\ref{eq:softmax_formulation}) is set to 1. The one-hot encoding $\hat{e}^c$ can be generated by taking the {\it argmax} of $\bar{e}^c$,
which is then mapped into a semantic feature vector $f_e^s \in \mathbb{R}^{128}$ with a single fully connected layer. 

While the semantic branch would be, in principle, sufficient to convey the entity identity to the remainder of the network, this does not suffice to infer visual relationships. For example, the detection of the ``people" and ``bike" entities in Figure~\ref{fig:model} is not enough to infer whether the relationship is ``person-standing by-bike" or ``person-riding-bike". This problem is addressed by introducing the appearance branch $F^a(\cdot;\phi)$ of parameter $\phi$, which outputs a feature vector $f_e^a \in \mathbb{R}^{128}$ with no pre-defined semantics, simply encoding entity appearance. Finally, the feature vectors $f_e^a$ and $f_e^s$ are  concatenated into a vector $f_e^{\{a,s\}}\in \mathbb{R}^{256}$ that represents both the appearance and semantics of entity $e$. 

\textbf{Predicate classifier} takes the subject $f_s^{\{a,s\}}$ and object $f_o^{\{a,s\}}$ feature vectors as input. These vectors are then concatenated with an embedding of subject $s^b$ and object $o^b$ bounding boxes produced by a fully-connected layer, to create a joint encoding $f_{\{s,o\}}^{\{a,s,b\}}\in \mathbb{R}^{520}$ of the semantics, appearance, and location of the subject-object patches $I^s$ and $I^o$. The predicate classifier $H$ is implemented with a small feature extractor $H(., \psi)$, consisting of three layers that perform dimension reduction. The input $f_{\{s,o\}}^{\{a,s,b\}}\in \mathbb{R}^{520}$ is first transformed into a $256$-dimension vector with a fully connected layer, followed by a batch normalization and a ReLU layer, the output of which is finally passed through a fully connected layer with a tanh non-linearity, to produce a final feature vector $f_{s,o} \in \mathbb{R}^{128}$. This is fed to a softmax layer to produce 
the probability 
of the predicate class
\begin{equation}
   \bar{p} = \rho(f_{s,o}, {\bf W}^p,\tau=1)
   \label{eq:softmax2}
\end{equation}
where ${\bf W}^p$ is the weight matrix of the predicate classifier.


\subsection{Training}\label{sec:training}
DT2 is trained with standard cross-entropy losses targeted on entity and predicate classification. The former is defined as
\begin{equation}
    L_{ent} = \frac{1}{n}\sum_{i=1}^n\frac{1}{|E_i|}\sum_{e_k\in E_i} L_{ce}(e_k^c, \bar{e}_k^c)
    \label{eq:objloss}
\end{equation}
where $L_{ce}$ denotes the cross-entropy loss, $\bar{e}_k^c$ is the output probability prediction of~(\ref{eq:softmax}) and $e_k^c$ is the ground truth one-hot code of the $k^{th}$ entity in the set $E_i$ from image $I_i$.
This is complemented by a predicate classification loss 
\begin{equation}
    L_{pred} = \frac{1}{n}\sum_{i=1}^n\frac{1}{|R_i|}\sum_{\substack{r_k=(s_k,p_k,o_k)\in R_i}}L_{ce}(p_k, \bar{p}_k)
    \label{eq:predloss}
\end{equation}
where $\bar{p}_k$ is the output probability of~(\ref{eq:softmax2}) and $p_k$ the ground truth one-hot code for the $k^{th}$ predicate in the set $R_i$ of visual relations in image $I_i$.
Both (\ref{eq:objloss}) and (\ref{eq:predloss}) are important to guarantee that the network can learn from both entities and predicate relationships.

\subsection{Sampling strategies}
While encapsulating both semantics and appearance information, the proposed training loss in Sec.~\ref{sec:training} requires a complementary sampling strategy tailored for long-tailed data. This long-tailed problem has been  studied mostly in the object recognition literature, where an image patch is fed to a feature extractor with the parameter $\varphi$ and the softmax layer $\rho$ of (\ref{eq:softmax_formulation}) with weight matrix ${\bf W}$. A popular training strategy is to use different sampling strategies to train the two network components~\cite{iclr2020}. The intuition is that, because the bulk of the network parameters are in the feature extractor ($\varphi$), this should be learned with the largest possible amount of data. Hence, the entire network is first trained with \textbf{Standard Random Sampling (SRS)}, which samples images uniformly, independent of their class labels. 

While this produces a good feature extractor, the resulting classifier usually overfits to the head classes, which are represented by many more images and have a larger weight on the cost function. The problem is addressed by fine-tuning the network on a balanced distribution, obtained with \textbf{Class Balanced Sampling (CBS)}. This consists of sampling uniformly over classes, rather than images, and guarantees that all classes are represented with equal frequencies. 
However, because images from tail classes are resampled more frequently than those of head classes, it carries some risk of overfitting to the former. To avoid overfitting, the fine-tuning is restricted to the weights $\bf W$ of the softmax layer. In summary, the network is trained in two stages. First, the parameters $\varphi$ and $\bf W$ are jointly learned with SRS. Second, the feature extractor ($\varphi$) is fixed and the softmax layer parameters $\bf W$ are relearned with CBS.

\subsection{Sampling for visual relationships}
Similar to long-tailed object recognition, it is sensible to train a model for visual relations in two stages. In the first stage, the goal is to learn the parameters $\theta, \phi, \psi$ of the feature extractors (see Sec.~\ref{sec:model}), which are the overwhelming majority of the network parameters. As in object recognition, the network should be trained with SRS. 
In the second stage, the goal is to fine-tune the softmax parameters ${\bf W}^e$ and ${\bf W}^p$ to avoid overfitting to head classes. 
However, unlike long-tailed object recognition, Figure~\ref{fig:clsR_baseline} shows that predicates and entities can have very different distributions, which makes the learning of long-tailed visual relations a distinct problem. 
This indicates that two class-balanced sampling strategies are required to accommodate the distribution difference between predicate and entity classes.

\begin{figure}
    \centering
    \includegraphics[width=0.85\linewidth]{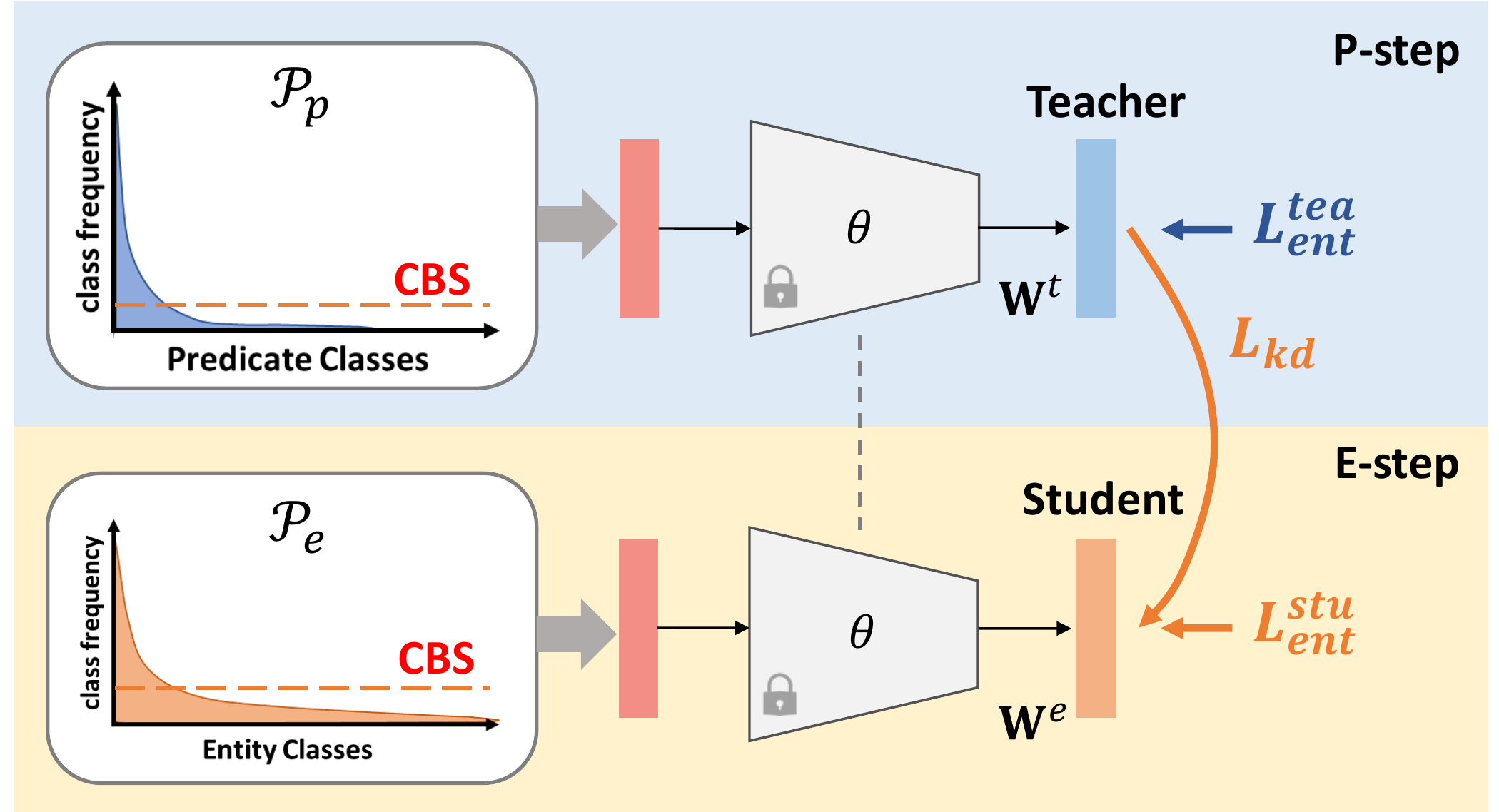}
    \caption{ACBS captures the interplay between the long-tailed distributions of entities and relations by implementing the knowledge distillation between P-step and E-step.}
    \label{fig:acbs}
\end{figure}
A straightforward solution is to introduce a 2-step iterative training procedure, namely \emph{entity-optimization step} (E-step) and \emph{predicate-optimization step} (P-step), to optimize the weight of ${\bf W}^e$ and ${\bf W}^p$ respectively. In E-step, images are sampled from a distribution ${\cal P}_e$ that is uniform with respect to entity classes, which is denoted as Entity-CBS. While in P-step, they are sampled from a distribution ${\cal P}_p$ uniform with respect to predicate classes, denoted as Predicate-CBS. However, since the uniform sampling of ${\cal P}_p$ is not class-balanced for entity classes, P-step would lead to the overfitting of the entity classification parameters ${\bf W}^e$. 

To address this problem, we propose a novel sampling strategy, \textbf{Alternating CBS} (ACBS), tailored for long-tailed visual relations. ACBS contains
a memory mechanism to maintain the entity predictions of the P-step, making sure that what was learned is not forgotten in the E-step. 
It is implemented with distillation~\cite{HinVin15Distilling} between the P-step and E-step and an auxiliary \textit{teacher} entity classifier of weight matrix ${\bf W}^t$. The \textit{teacher} entity classifier is inserted in parallel with the entity classifier of weight matrix ${\bf W}^e$ in (\ref{eq:softmax}), which is its \textit{student}, and produces a second set of 
entity prediction probabilities as

\begin{align}
    \bar{e}^t = \rho(F^s(f_e;\theta), {\bf W}^t, \tau=1). \label{eq:sofmax_teacher}
\end{align}
With the introduction of the teacher entity classifier, we rewrite (\ref{eq:objloss}) into $L_{ent}^{stu}$ and $L_{ent}^{tea}$, where the former operates on $\bar{e}^c$ of (\ref{eq:softmax}) and the latter operates on $\bar{e}^t$. 
Furthermore, to distill knowledge from the teacher entity classifier, a Kullback-Leibler divergence (KL) loss ($L_{kd}$) is defined as
\begin{equation}
     \text{KL}({\rho(F^s(f_e;\theta), {\bf W}^e, \tau=\tau_s) || \rho(F^s(f_e;\theta), {\bf W}^t, \tau=\tau_s)}),
      \label{eq:kdloss}
\end{equation}
where the two inputs to $L_{kd}$ are the smooth version of (\ref{eq:softmax}) and (\ref{eq:sofmax_teacher}) with temperature $\tau_s$.

In summary, the P-step updates parameters ${\bf W}^p$ of the predicate classifier and ${\bf W}^t$ of the teacher with (\ref{eq:predloss}) and $L_{ent}^{tea}$ respectively, while the student parameters ${\bf W}^e$ are kept fixed. In the E-step, ${\bf W}^p$ and ${\bf W}^t$ (teacher) are kept fixed, and ${\bf W}^e$ (student) is optimized with $L_{ent}^{stu}$ and (\ref{eq:kdloss}). 
This implements learning without forgetting~\cite{lwf2016} between the two steps, encouraging the student classifier to mimic the predictions of the teacher classifier, and enabling the network to learn the new parameters for one distribution, e.g. ${\bf W}^e$, without forgetting the one, e.g. ${\bf W}^t$, previously learned for the other. The training procedure is detailed in Algorithm~\ref{algo:ACBS}.

\begin{algorithm}[t]
\SetAlgoLined
\textbf{Input:} Training dataset $\mathcal{D}$, predicate distribution ${\cal P}_p$, entity distribution ${\cal P}_e$, ACBS hyperparameters ($\alpha,\beta,\tau_s$), and model parameters ($\theta,\phi,\psi$).

\textbf{Output:} Model parameters ($\mathbf{W}^p$, $\mathbf{W}^e$).  

 \While{Not convergence}{
 \tcp{P-Step}
 $\mathcal{D}_p \leftarrow$ \textit{BalancedSample}($\mathcal{D}$, ${\cal P}_p$)\;
 \While{$batch$ in  $\mathcal{D}_p$}{
    $L_{total} \leftarrow L_{pred}~(\ref{eq:predloss}) + \beta L_{ent}^{tea}$~(\ref{eq:objloss})\;
    
    \text{Minimize $L_{total}$ with respect to $(\mathbf{W}^p,\mathbf{W}^t)$}
 }
 
 \tcp{E-Step}
 $\mathcal{D}_e \leftarrow$ \textit{BalancedSample}($\mathcal{D}$, ${\cal P}_e$)\;
 \While{$batch$ in  $\mathcal{D}_e$}{
    $L_{total} \leftarrow L_{ent}^{stu}~(\ref{eq:objloss}) + \alpha L_{kd}$~(\ref{eq:kdloss})\;
    
    \text{Minimize $L_{total}$ with respect to $\mathbf{W}^e$}
 }
 \caption{Training procedure of ACBS}\label{algo:ACBS}
}
 \vspace{-0mm}
\end{algorithm}

\section{Experiments} 
\label{sec:experiment}
In  this  section,  several experiments are performed to validate the effectiveness of DT2-ACBS.

\begin{table*}[t!]
\centering
\caption{The result (mRecall@K) of SGG tasks (PredCls, SGCls, SGDet) compared to SOTA in scene graphs. Results for other methods are reported from the corresponding paper in general. $\dagger$ denotes our reproduced model with ResNet101-FPN backbone. 
}
\setlength{\tabcolsep}{9pt}
\resizebox{1\linewidth}{!}{
\begin{tabular}{c|ccc|ccc|ccc}
\hline
  & \multicolumn{3}{|c}{Predicate Classification} & \multicolumn{3}{|c}{Scene Graph Classification} & \multicolumn{3}{|c}{Scene Graph Detection}\\
Method & mR@20 & mR@50 & mR@100 & mR@20 & mR@50 & mR@100 & mR@20 & mR@50 & mR@100\\ \hline\hline
IMP+~\cite{xu2017scenegraph}         & -    & 9.8  & 10.5 & -    & 5.8 & 6.0      &- & 3.8&4.4\\
FREQ~\cite{neural_motifs_2017}       & 8.3  & 13.0 & 16.0 & 5.1  & 7.2 & 8.5      &4.5&6.1&7.1\\
MOTIFS~\cite{neural_motifs_2017}     & 10.8 & 14.0 & 15.3 & 6.3  & 7.7 & 8.2      &4.2&5.7&6.6\\
MOTIFS~\cite{neural_motifs_2017}$\dagger$   & 13.2 & 16.3 & 17.5 & 7.1& 8.8 & 9.3     &4.9&6.7&8.2\\
KERN~\cite{kern_CVPR19}              & -    & 17.7 & 19.2 & -    & 9.4 & 10.0     &-&6.4&7.3\\
VCTree~\cite{VCTree_Tang_2019_CVPR}  & 14.0 & 17.9 & 19.4 & 8.2  & 10.1 & 10.8    &5.2&6.9&8.0\\
GBNet~\cite{Zareian_2020_ECCV}       & -    & 22.1 & 24.0 & -    & 12.7 & 13.4    &-&7.1&8.5\\
TDE-MOTIFS-SUM~\cite{TDETangNHSZ20}  & 18.5 & 25.5 & 29.1 & 9.8  & 13.1 & 14.9    &5.8&8.2&9.8\\
TDE-MOTIFS-SUM~\cite{TDETangNHSZ20}$\dagger$  & 17.9 & 24.8 & 28.6 & 9.6& 13.0& 14.7   &5.6&7.7&9.1\\
TDE-VCTree-SUM~\cite{TDETangNHSZ20} & 18.4 & 25.4 & 28.7 & 8.9 & 12.2 & 14.0    &6.9&9.3&11.1\\
TDE-VCTree-GATE~\cite{TDETangNHSZ20} & 17.2 & 23.3 & 26.6 & 8.9 & 11.8 & 13.4    &6.3&8.6&10.3\\
PCPL~\cite{PCPL_YanSJHJC020}         & -    & 35.2 & 37.8 & -    & 18.6 & 19.6    &-&9.5& 11.7\\\hline\hline
DT2-ACBS (ours) & \textbf{27.4} & \textbf{35.9} & \textbf{39.7} & \textbf{18.7} & \textbf{24.8} & \textbf{27.5} &\textbf{16.7} & \textbf{22.0} & \textbf{24.4} \\\hline
\end{tabular}\label{tab:main_table}
}
\end{table*}

\begin{table*}[t!]
\centering
\caption{mR@100 on SGG tasks for head, middle, tail classes. $\dagger$ denotes our reproduced models with ResNet101-FPN backbone.}\label{tab:3class}
\setlength{\tabcolsep}{9pt}
\resizebox{1\linewidth}{!}{
\begin{tabular}{c|ccc|ccc|ccc}
\hline
  & \multicolumn{3}{|c}{Predicate Classification} & \multicolumn{3}{|c}{Scene Graph Classification} & \multicolumn{3}{|c}{Scene Graph Detection}\\
Method & Head (16) & Middle (17) & Tail (17) & Head (16) & Middle (17) & Tail  (17) & Head (16) & Middle (17) & Tail (17)\\ \hline\hline
MOTIFS~\cite{neural_motifs_2017}$\dagger$    & 42.3 & 9.8 & 0.6 & 24.6 & 4.0 & 0.1 & 20.2 & 4.6 & 0.4\\
TDE-MOTIFS-SUM~\cite{TDETangNHSZ20}$\dagger$   & \textbf{44.9} & 35.8 & 6.1 & \textbf{25.6} & 15.8 & 3.3 & 22.2 & 5.6 & 0.1\\ \hline\hline
DT2-ACBS (ours) & 35.1 & \textbf{45.2} & \textbf{38.6} & 24.6 & \textbf{29.1} & \textbf{28.6} & \textbf{22.3} & \textbf{26.7} & \textbf{24.0}\\\hline
\end{tabular}
}
\end{table*}

\subsection{Dataset}
Visual Genome (VG)~\cite{visual_genome16} is composed of 108k images across 75k object categories and 37k predicate categories, but $92\%$ of the predicates have less than 10 instances. Following prior works, we use the original splits of the popular subset (i.e. VG150) for training and evaluation. It contains the most frequent 150 object classes and 50 predicate classes. The distribution remains highly long-tailed. To perform balanced sampling during training, predicate classes with less than 5 instances, e.g. ``flying in," are ignored.

\begin{figure}[t]
    \centering
    \includegraphics[width=0.46\textwidth,height=0.5\linewidth]{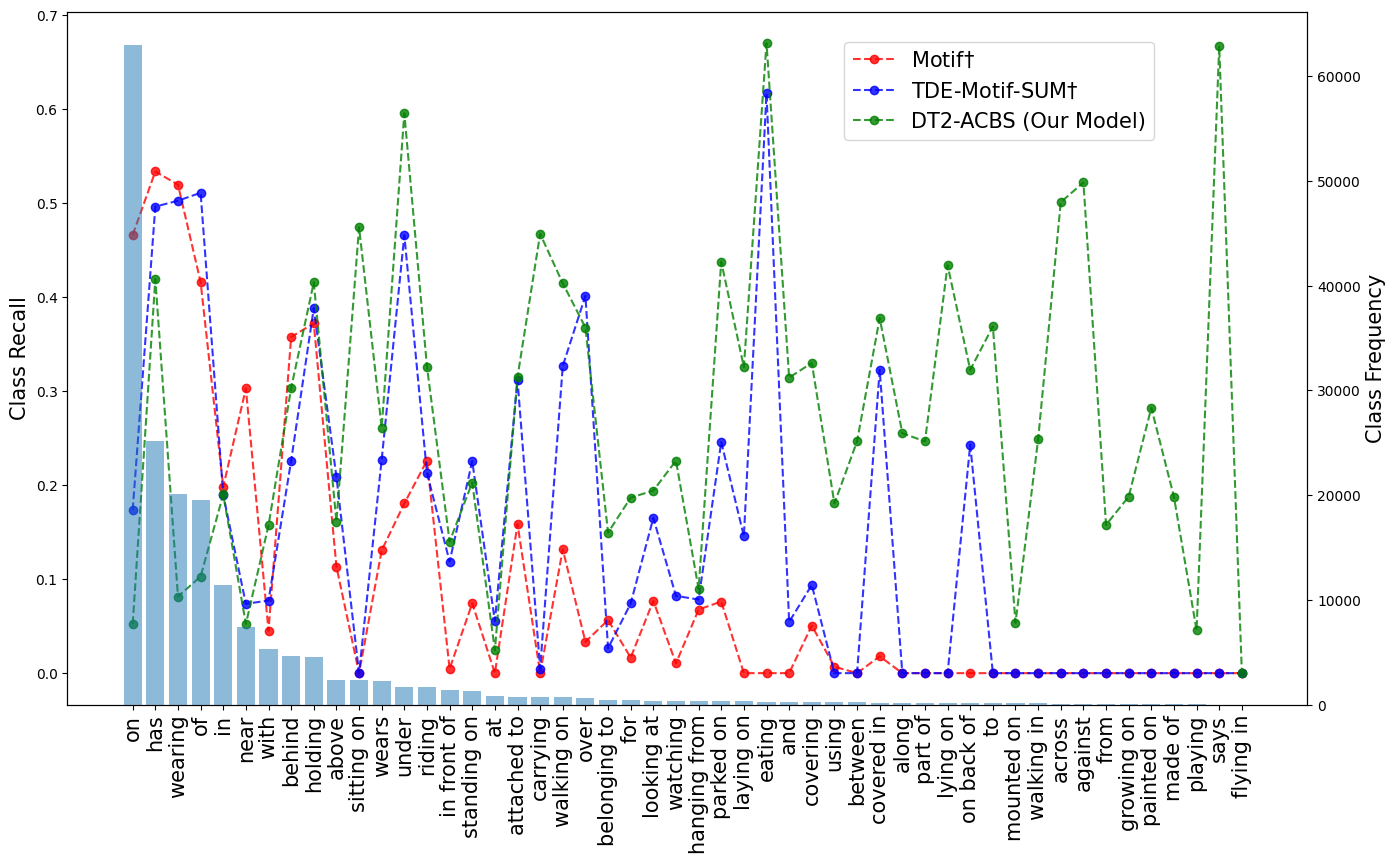}
    \caption{Comparisons of per class Recall@100 on SGCls.
    Classes are sorted in decreasing order of the number of samples.}
    \label{fig:sgcls_histogram}
\end{figure}

\subsection{Comparison to SOTA}
To validate our hypothesis, we compare DT2-ACBS with the state-of-the-art methods on PredCls, SGCls and SGDet task on the popular subset VG150 of VG~\cite{visual_genome16}, under the mRecall@K metric. As shown in Table~\ref{tab:main_table}, compared baselines include 1) simple frequency-based method~\cite{neural_motifs_2017}, 2) sophisticated architecture design for contextual representation learning~\cite{xu2017scenegraph,kern_CVPR19,VCTree_Tang_2019_CVPR,Zareian_2020_ECCV} and 3) recent works that tackle the long-tailed bias of predicate classes~\cite{TDETangNHSZ20,PCPL_YanSJHJC020}.
Several observations can be made.
First, DT2-ACBS outperforms all baselines in the first two groups by a large margin (mR@100 gain larger than $15.7\%$) on the PredCls task, where entity bounding boxes and categories are given. 
The baselines in the third group~\cite{TDETangNHSZ20,PCPL_YanSJHJC020}, which address the long-tailed bias of the predicate distribution, are similar in spirit to DT2-ACBS. However, the latter relies on a simpler model design and a more sophisticated decoupled training scheme to overcome overfitting.
This enables a $1.9\%$ improvement on mR@100 ($5\%$ relative improvement), showing the efficacy of the proposed sampling mechanism for tackling the long-tailed problem in predicates distribution. 

Next, when predicting both predicate and entity class given the ground truth bounding boxes (SGCls task), DT2-ACBS outperforms all existing methods by a larger mR@100 margin ($1.9\%$ on PredCls vs $7.9\%$ on SGCls, equivalently relative improvement of 5\% in PredCls vs 40\% in SGCls). This significant improvement in SGCls performance can be ascribed to the decoupled training of ACBS, which better captures the interplay between the different distributions of entities and predicates. 

Finally, we also ran DT2-ACBS on proposal boxes generated by a pre-trained Faster-RCNN for the SGDet task.
Table~\ref{tab:main_table} shows that DT2-ACBS 
outperforms existing methods by a significantly larger mR@100 margin of $12.7\%$ ($>100\%$ relative improvement) on the SGDet task. 

\begin{figure*}[t]
\begin{tabular}{cccc}
\centering
\includegraphics[width=0.5\linewidth]{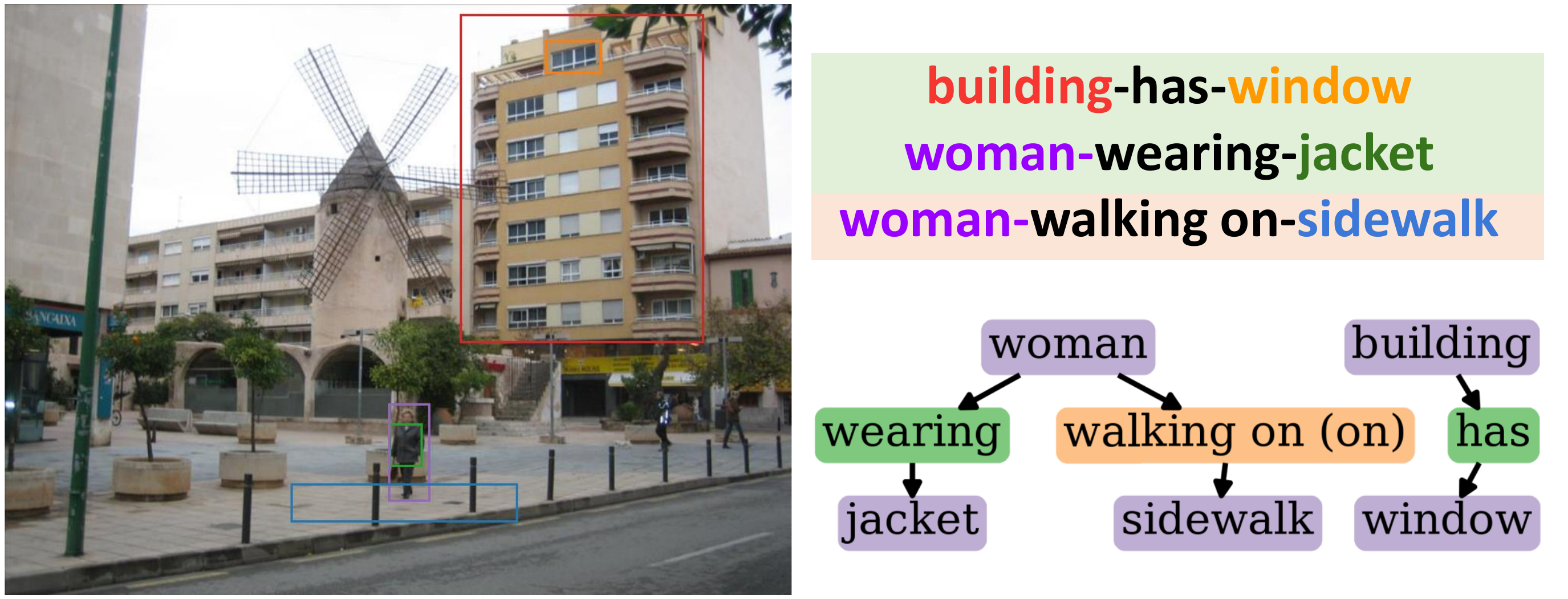}
& \includegraphics[width=0.45\linewidth]{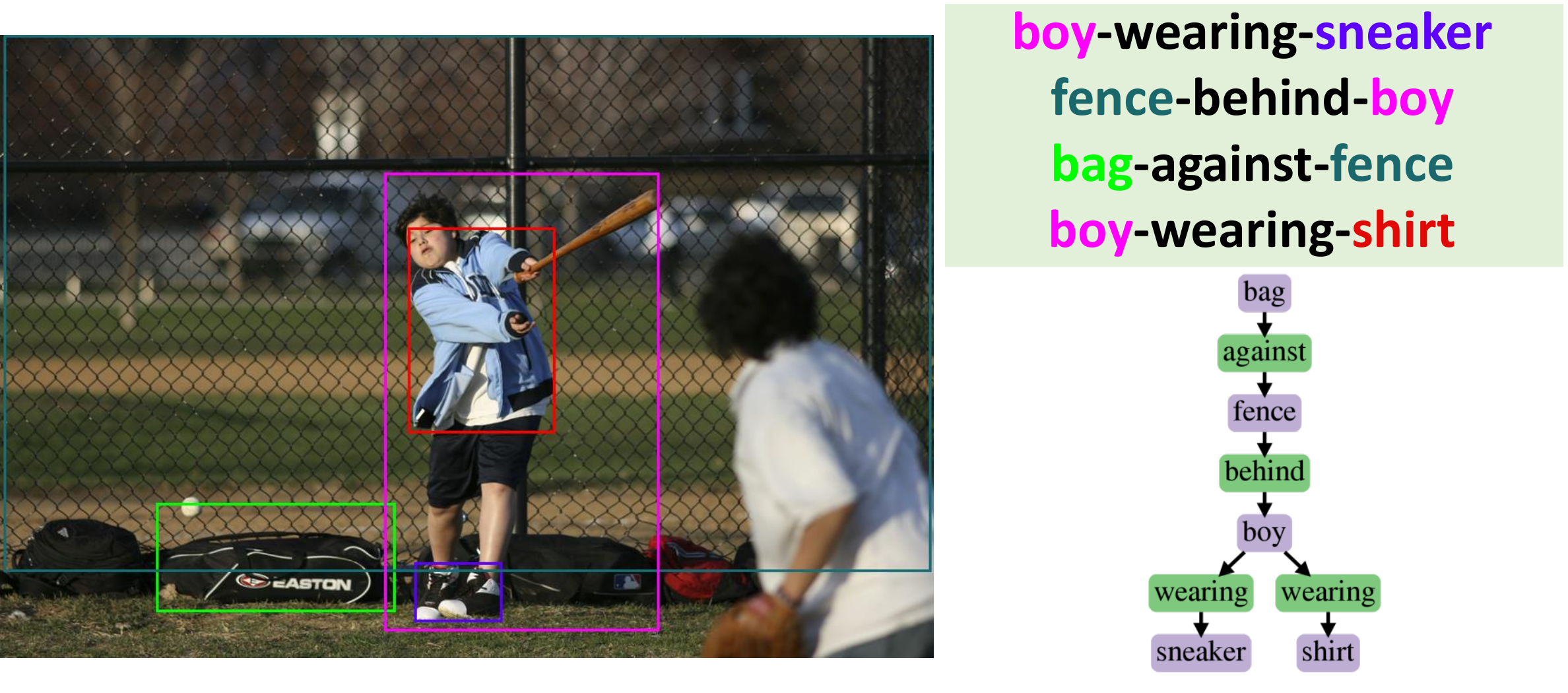}
\end{tabular}
\caption{\textbf{Qualitative results of PredCls (left) and SGCls (right).} In each sub-figure, 
colors of bounding boxes in the image (left) are corresponding to the entities in the triplets (upper-right) with the background color green/orange for correct/incorrect predicate predictions. In the generated graphs (lower-right), correct/incorrect predictions of entities and predicates are shown in purple/blue and green/orange respectively, with the ground truth noted in the bracket (best viewed in color). More examples are shown in the supplemental.
}  
\label{fig:qualitative_results}
\end{figure*}


\noindent\textbf{Class-wise performance analysis:}
To study the performance of classes with different popularity, we sort the 50 relation classes by their frequencies and divide them into 3 equal parts, head (16), middle (17) and tail (17). Table~\ref{tab:3class} presents the mR@100 performance on these partitions for each SGG task. As observed in prior long-tailed recognition work~\cite{oltr,iclr2020}, a performance drop in head classes is hard to avoid while improving tail class performance.
The goal, instead, is to achieve the best balance among all the classes, which DT2-ACBS clearly does with notable improvements in the middle and tail classes. 
It should also be noted that the drop in head performance can be deceiving, due to dataset construction problems like ``wearing" and ``wears" appearing as two different relationship classes. Most importantly, many VG150 tail categories (e.g. ``standing on", ``sitting on") are fine-grained versions of
a head category (``on"). Some of the degradation in head class performance is just due to the predicates being pushed to the fine-grained classes, which is more informative. 
We notice that one of the high-frequency predicate classes \emph{On} has a low recall value (Figure~\ref{fig:sgcls_histogram}) and observe that DT2-ACBS often instead predicts its fine-grained sub-categories, such as \emph{standing on}, \emph{sitting on}, \emph{mounted on}.
In particular, there are $41,620$ ground truth instances of \emph{On} predicate in the test set, and DT2-ACBS predicts \emph{On}-subcategories $14,317$ times on PredCls, which constitutes $34\%$ incorrect predictions as per the metric. 
Overall, DT2-ACBS performs significantly better in middle and tail classes on SGG tasks, and performs comparably on head classes for SGCls and SGDet, reaching the best balance across all the classes. 

\begin{table}[t!]
\centering
\caption{Ablations on different sampling strategies for SGCls.
}
\resizebox{.9\linewidth}{!}{
\setlength{\tabcolsep}{7pt}
\begin{tabular}{l|ccc}
\hline
Method & mR@20 & mR@50 & mR@100  \\\hline\hline
Single Stage-SRS & 6.4 & 9.6 & 11.2 \\
Single Stage-Indep. CBS & 8.5 & 11.2 & 12.4 \\ \hline\hline
DT2-Predicate-CBS & 10.0 & 13.0 & 14.3 \\
DT2-Indep. CBS & 17.3 & 23.9 & 26.7 \\
DT2-ACBS (ours) & \textbf{18.7} & \textbf{24.8} & \textbf{27.5} \\
\hline
\end{tabular}
}
\label{tab:ablation_SGCls}
\end{table}

\subsection{Ablations on sampling strategies}
SGCls performance is affected by the intertwined entity and predicate distributions. In this section, we conduct ablation studies in Table~\ref{tab:ablation_SGCls} on 1) single-stage vs two-stage training and 2) different sampling schemes. The first half of the table shows the performances of single-stage training, where the representation and the classifier are learned together. This clearly under-performs the two-stage training, which is listed in the second half of the table, where we compare different sampling strategies in the second stage of DT2. For the predicate classifier, it can be trained based on either SRS or class-balanced sampling for predicates (Predicate-CBS). Since each relation comes with a subject and an object, it is possible to train the entity classifier with respect to Predicate-CBS, indicating the entity classifier can be trained based on SRS, Predicate-CBS or class-balanced sampling for entities (Entity-CBS). Note that the predicate classifier can not be trained with Entity-CBS, since an entity does not always belong to a visual relation tuple. From the second half of the table, we find that considering the distribution differences in predicates and entities is important, because DT2-Predicate CBS (i.e. Predicate-CBS for both entity and predicate classifier) does not perform as well as DT2-Indep. CBS (i.e. Entity-CBS for the entity classifier and Predicate-CBS for the predicate classifier).
The observation that DT2-Indep. CBS already performs better than existing methods (Table \ref{tab:main_table}) supports our claim that visual relations can be effectively modeled with a simple architecture if the long-tailed aspect of the problem is considered. Nevertheless, the proposed ACBS further improves the SGCls performance by distilling the knowledge between P-step and E-step (see Algorithm \ref{algo:ACBS}).

\subsection{Qualitative results}\label{sec:visual_ex}
Figure \ref{fig:qualitative_results} presents qualitative results of DT2-ACBS. In PredCls task, DT2-ACBS can correctly predict populated predicate classes (\emph{has} \& \emph{wearing}) as well as non-populated predicate classes (\emph{walking on}). 
Not only robust to long-tailed predicate classes, DT2-ACBS is also able to classify entities ranging from more populated classes (\emph{boy}) to tail classes (\emph{sneaker}). 
We can observe that while the predicted predicates can be different from the ground truth, the relation can still be reasonable (e.g. a \textit{subclass} or a \textit{synonym} of the ground truth). For example, the predicted predicate ``walking on" is actually a subclass of the ground truth predicate ``on”. These examples show that DT2-ACBS is able to predict more fine-grained predicates in tail classes and provide more exciting descriptions of the scene.

\section{Conclusions}
\label{sec:conclusions}
Learning visual relations is inherently a long-tailed problem. Existing approaches have mostly proposed complex models to learn visual relations. However, complex models are ill-suited for long-tailed problems, due to their tendency to overfit. 
In this paper, we consider the uniqueness of visual relations, where entities and relations have skewed distributions. We propose a simple model, namely DT2,
along with an alternating sampling strategy (ACBS) to tackle the long-tailed visual relation problem.
Extensive experiments on the benchmark VG150
dataset show that DT2-ACBS significantly outperforms the state-of-the-art methods of more complex architectures. 

\noindent\textbf{Acknowledgements}
This work was funded by NSF awards IIS-1924937, IIS-2041009 and a gift from Amazon.

{\small
\bibliographystyle{ieee_fullname}

}

\clearpage
\appendix

\section{Model complexity}
The model complexity is quite low for DT2, which has \textbf{10$\times$} fewer trainable parameters than most of the recent approaches in the literature. For example, the SGCls model sizes of DT2, MOTIFS~\cite{neural_motifs_2017}, VCTree~\cite{VCTree_Tang_2019_CVPR} and TDE-MOTIFS~\cite{TDETangNHSZ20} are \textbf{224 MB}, 1.68 GB, 1.65 GB and 2.1 GB respectively. 
This is by design, since our goal is to emphasize the importance of accounting for long tails during training.

\begin{figure*}[t!]
\centering
  \begin{tabular}{cc}
      \includegraphics[width=0.5\linewidth]{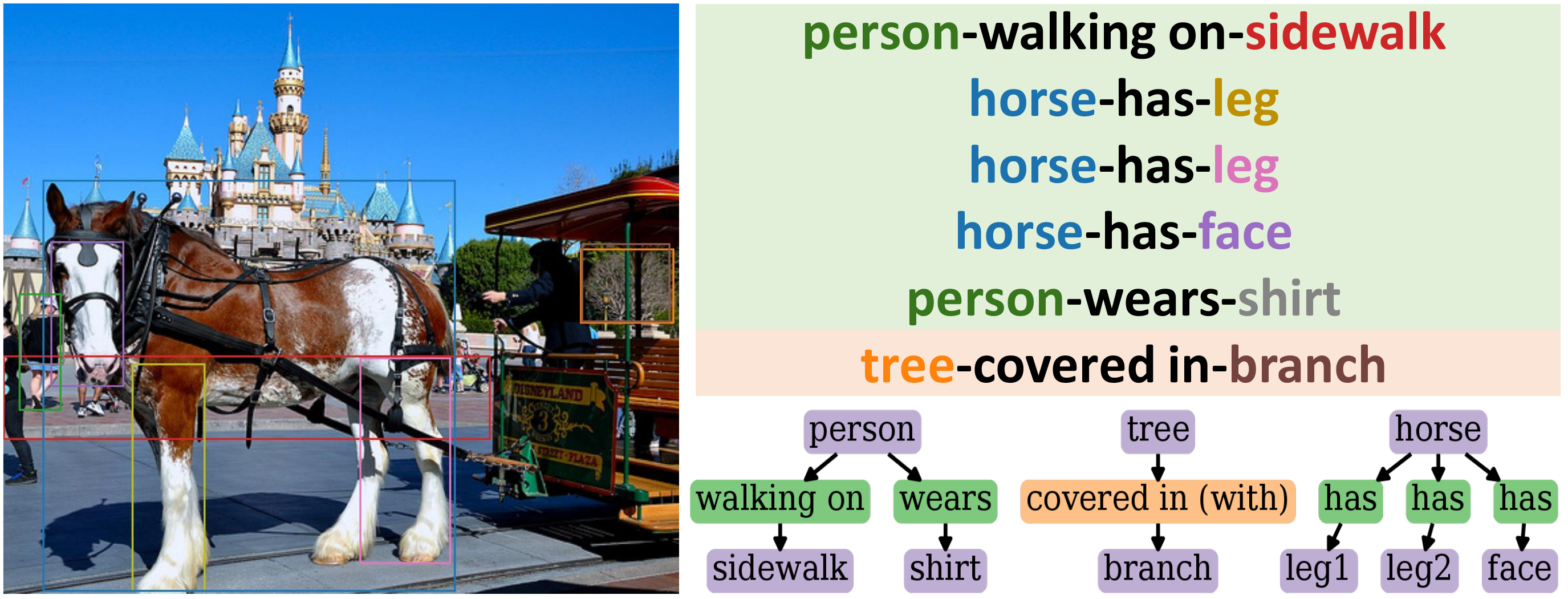} &
      \includegraphics[width=0.42\linewidth]{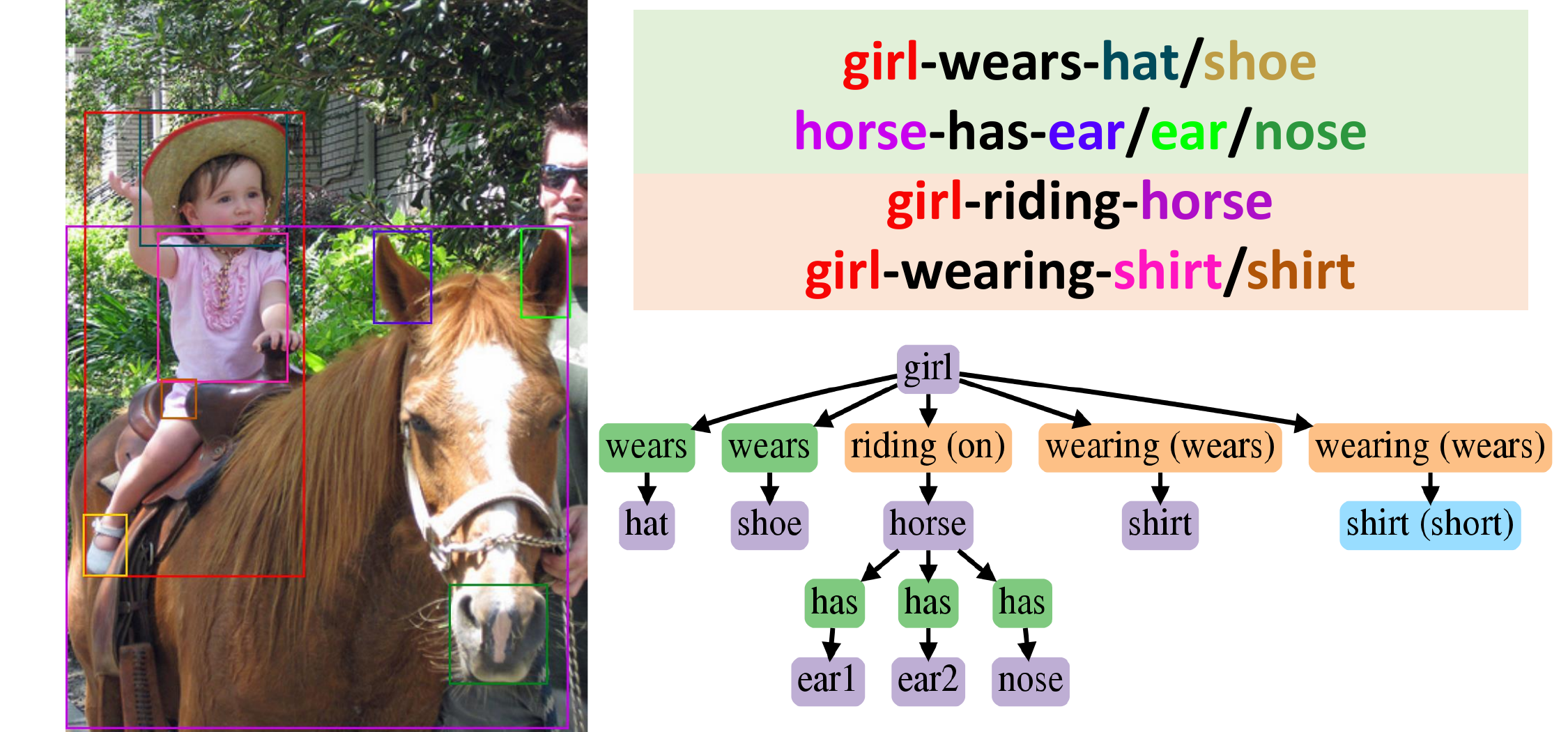}\\
      \includegraphics[width=0.5\linewidth]{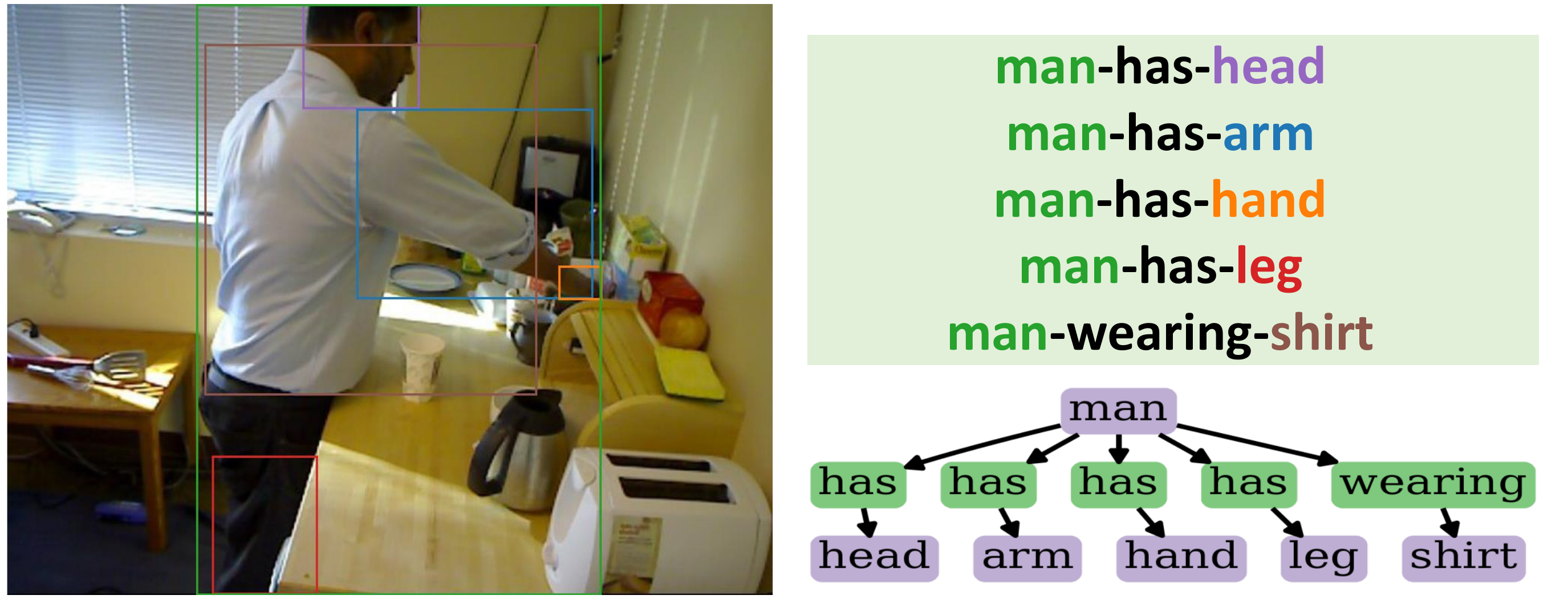} &
      \includegraphics[width=0.45\linewidth]{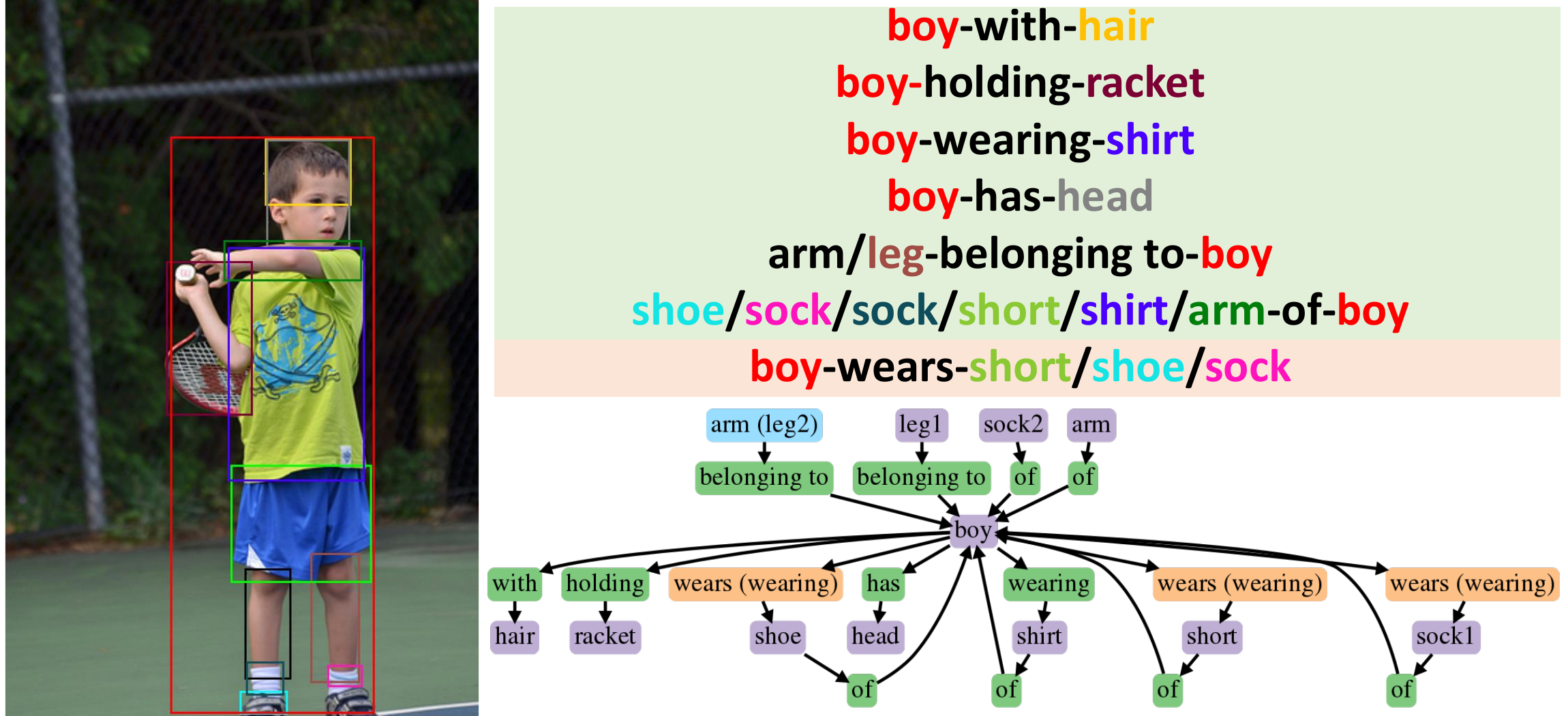} \\
      \includegraphics[width=0.5\linewidth]{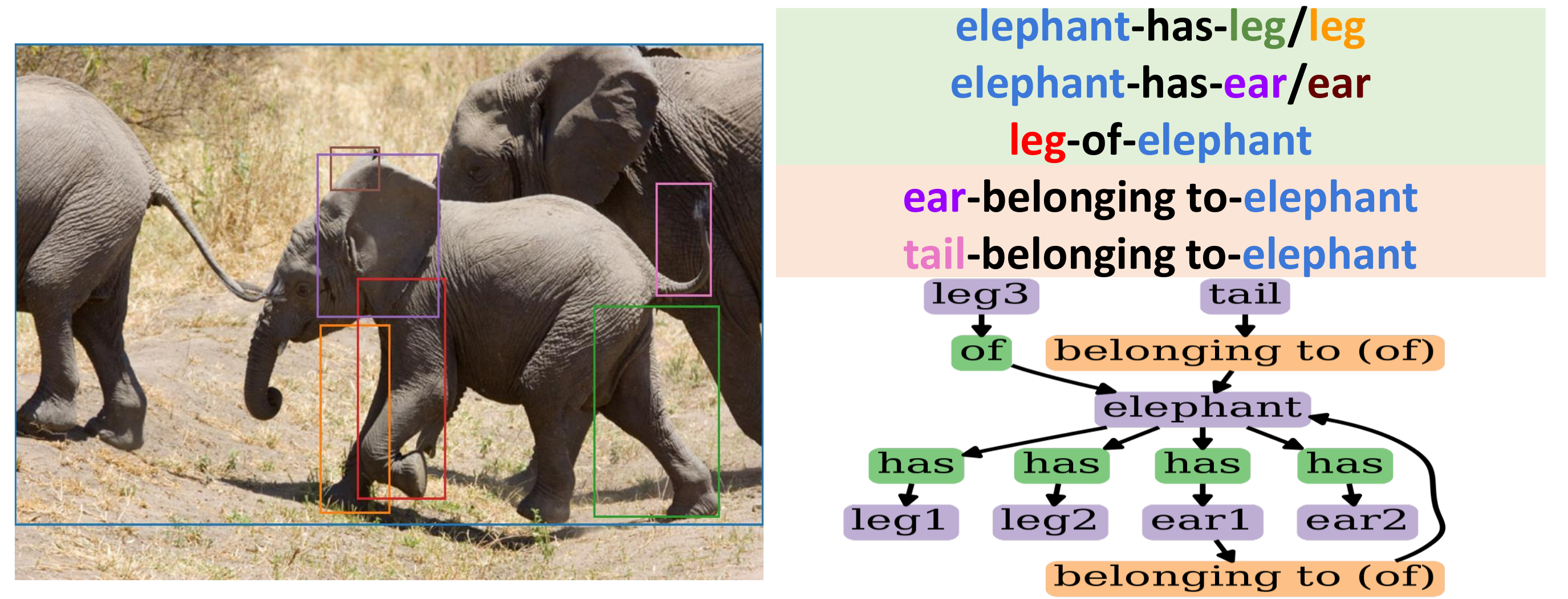} & 
      \includegraphics[width=0.45\linewidth]{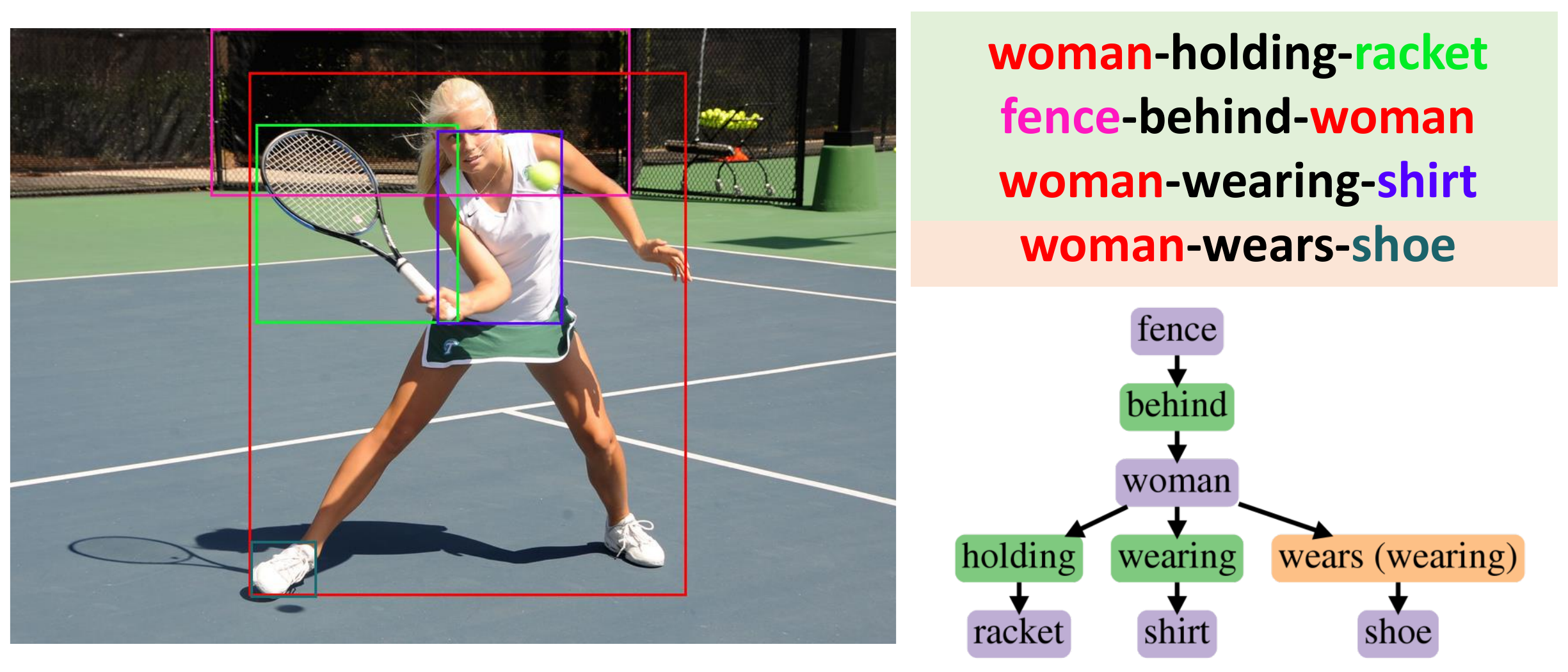}
  \end{tabular}
  \caption{\textbf{Additional visual examples of PredCls (left column) and SGCls (right column).} In each sub-figure, colors of bounding boxes in the image (left) are corresponding to the entities in the triplets (upper-right) with the background color green/orange for correct/incorrect predicate predictions. In the generated graphs (lower-right), correct/incorrect predictions of entities and predicates are shown in purple/blue and green/orange respectively, with the ground truth noted in the bracket (best viewed in color).
}\label{fig:example}  
\end{figure*}

\section{Additional results}
In this section, we provide additional results of the proposed DT2-ACBS.
\subsection{More visual examples}
In addition to section~\ref{sec:visual_ex}, Figure \ref{fig:example} presents more visual examples of PredCls (left column) and SGCls (right column) generated by DT2-ACBS. These examples show that DT2-ACBS can predict predicates ranging from head classes such as {\it has} and {\it wearing} to less populated classes like {\it walking on}. While some examples are counted as incorrect predictions under the metric, as discussed in the main paper, they are still reasonable predictions such as a subclass or a synonym of the ground truth. For example, {\it belonging to/of} and {\it wears/wearing}. In the SGCls task, DT2-ACBS correctly predicts head entities ({\it boy}, {\it horse}, and {\it head}) and tail entities ({\it racket} and {\it sock}). 

\subsection{Ablations on appearance branch}
As discussed in section~\ref{sec:model}, the goal of appearance branch is to convey the image information {\it not\/} encoded in the entity labels but {\it relevant to predicate predictions.\/} 
We tested the effectiveness of the appearance branch $F^a$ by removing it and training the network with ACBS. Table~\ref{table:appearance} shows that entity classification accuracy remains similar, but PredCls and SGCls performance drops dramatically, i.e. the appearance branch contributes substantially to predicate classification. Note that the gains hold even when the ground truth entity labels are used (PredCls), confirming the argument that simply knowing entity classes is not enough for predicate prediction.

\subsection{E-step as teacher}
Note that the predicate ${\bf W}^p$ and entity ${\bf W}^e$ weight matrices are interdependent. Using E-step as the teacher would negatively affect ${\bf W}^p$. In ACBS, ${\bf W}^e$ receives class-balanced {\it entity supervision,\/} so there is no risk of overfitting. The role of the teacher is to guarantee that the E-step update of ${\bf W}^e$ is not incompatible with the P-step update of ${\bf W}^p$. This distillation is exactly how ACBS fuses the knowledge learnt with different distributions. Using E-step as the teacher has weaker results, as shown in Table~\ref{table:teacher}.

\subsection{Recall values}
The metric of Recall@K is highly biased toward dominated classes (such as ``on"), and thus it is not suitable for long-tailed visual relations, as discussed in the main paper. However, we provide the numbers in Table~\ref{table:recall} for reference.

\begin{table}[t!]
\centering
\caption{Ablations of appearance branch in SGCls. (subj, obj) Acc. denotes the accuracy of a pair of subject and object class.}
\resizebox{\linewidth}{!}{
\begin{tabular}{c|c|c|c}
\hline
\multirow{2}{*}{Method} & PredCls & SGCls & (subj, obj) \\ 
& mR@ 20 / 50 / 100 & mR@ 20 / 50 / 100 & Acc. \\\hline\hline
w/o $F^a$      & 18.1 / 24.5 / 26.8 & 11.0 / 14.7 / 16.3 & 25.77 \\
w/ $F^a$ (ours) & \textbf{27.4} / \textbf{35.9} / \textbf{39.7} & \textbf{18.7} / \textbf{24.8} / \textbf{27.5} & 26.26 \\ \hline
\end{tabular}
}
\label{table:appearance}
\end{table}

\begin{table}[t!]
\centering
\caption{Ablations of ACBS with different teachers in SGCls.}
\resizebox{0.6\linewidth}{!}{
\setlength{\tabcolsep}{12pt}
\begin{tabular}{c|c}
\hline
Teacher & mR@ 20 / 50 / 100 \\ \hline\hline
E-step &  15.2 / 20.2 / 22.0 \\
P-step (ours) & \textbf{18.7} / \textbf{24.8} / \textbf{27.5} \\ \hline
\end{tabular}
}
\label{table:teacher}
\end{table}

\begin{table*}[t!]
\centering
\caption{Recall and mRecall values for SGG tasks.}
\tiny
\resizebox{\linewidth}{!}{
\setlength{\tabcolsep}{7pt}
\begin{tabular}{c|cc|cc|cc}
\hline
\multirow{2}{*}{Method} & \multicolumn{2}{c|}{Predicate Classification} & \multicolumn{2}{c|}{Scene Graph Classification} & \multicolumn{2}{c}{Scene Graph Detection} \\
& R@50 / 100 & mR@50 / 100 & R@50 / 100 & mR@50 / 100 & R@50 / 100 & mR@50 / 100 \\\hline\hline
KERN~\cite{kern_CVPR19} & 65.8 / 67.6 & 17.7 / 19.2 & 36.7 / 37.4 & 9.4 / 10.0 & 27.1 / 29.8 & 6.4 / 7.3 \\
TDE-MOTIFS-SUM~\cite{TDETangNHSZ20} & 46.2 / 51.4 & 25.5 / 29.1 & 27.7 / 29.9 & 13.1 / 14.9 & 16.9 / 20.3 & 8.2 / 9.8\\
TDE-VCTree-SUM~\cite{TDETangNHSZ20} & 47.2 / 51.6 & 25.4 / 28.7 & 25.4 / 27.9 & 12.2 / 14.0 & 19.4 / 23.2 & 9.3 / 11.1\\
PCPL~\cite{PCPL_YanSJHJC020}  & 50.8 / 52.6 & 35.2 / 37.8 & 27.6 / 28.4  & 18.6 / 19.6 &  14.6 / 18.6 & 9.5 / 11.7\\\hline\hline
DT2-ACBS (ours) & 23.3 / 25.6 & 35.9 / 39.7 & 16.2 / 17.6 & 24.8 / 27.5 & 15.0 / 16.3 & 22.0 / 24.4\\\hline
\end{tabular}
}
\label{table:recall}
\end{table*}

\section{Implementation Details}
DT2-ACBS is a two-stage training process. While SRS is adopted in the first stage when training the parameter of $\theta$, $\phi$ and $\psi$, the proposed ACBS is adopted in the second stage to learn the classifiers. 
Apart from the differences in sampling strategies, both stages share a similar optimization scheme, where the Adam optimizer with initial learning rate $10^{-3}$ is adopted, with the learning rate decay of $0.5$ for every 5 epochs. The batch size in the first stage is $256$, while in the second stage, objects and predicates are sampled with $2$ and $5$ samples per class respectively. The hyperparameters $\alpha$, $\beta$ and $\tau_s$ are set to  $0.2$, $1$ and $10$ respectively using the validation set. 
For evaluation on SGG tasks, we adopt the protocol of~\cite{neural_motifs_2017, VCTree_Tang_2019_CVPR} to filter out the subject-object pairs that do not have a relationship.
\end{document}